\documentclass[final,runningheads,a4paper]{llncs}

\usepackage{amssymb}
\usepackage{graphicx}
\usepackage{pgf}

\usepackage[obeyDraft]{todonotes}

\newcommand{\todoBP}[1]{\todo[author=BP,size=\small,color=blue!40]{#1}}

\newcommand{\hm}{\ensuremath H} %
\newcommand{\en}{\ensuremath E} %

\usepackage{url}
\urldef{\mailsa}\path|{alfred.hofmann, ursula.barth, ingrid.haas, frank.holzwarth,|
\urldef{\mailsb}\path|anna.kramer, leonie.kunz, christine.reiss, nicole.sator,|
\urldef{\mailsc}\path|erika.siebert-cole, peter.strasser, lncs}@springer.com|    
\newcommand{\keywords}[1]{\par\addvspace\baselineskip
\noindent\keywordname\enspace\ignorespaces#1}

\newcommand{\pdfTitle}{Femoral ROIs and Entropy for Texture-based Detection of Osteoarthritis from High-Resolution Knee Radiographs}
\newcommand{\pdfAuthor}{Hladuvka et al}
\newcommand{\pdfSubject}{}
\newcommand{\pdfKeywords}{computer-aided diagnosis, quantitative image analysis, X-ray imaging, imaging biomarkers}
\usepackage{color}
\definecolor{VRVis}{rgb}{0,0,0}

\usepackage{hyperref}
\hypersetup{                   
        colorlinks  = true,
        citecolor   = VRVis,
        linkcolor   = VRVis,   
        anchorcolor = VRVis,   
        filecolor   = VRVis,   
        menucolor   = VRVis,   
        pagecolor   = VRVis,   
        urlcolor    = VRVis,   
	bookmarksopen = true,
        pdfhighlight = /P,     
        pdfborder    = 0 0 0,
        pdftitle     = {\pdfTitle},       
        pdfauthor    = {\pdfAuthor},      
        pdfsubject   = {\pdfSubject},     
        pdfkeywords  = {\pdfKeywords},    %
        pdfcreator   = {},                
        pdfproducer  = {},                
        pdfpagemode  = UseOutlines,       
        pdfstartpage = 1,                 
	breaklinks   = true               
}

\begin{document}

\mainmatter  %

\title{%
Femoral ROIs and Entropy for
Texture-based Detection of Osteoarthritis from
High-Resolution
Knee Radiographs 
}
\titlerunning{
Femoral ROIs and entropy for texture-based detection of knee osteoarthritis
}

\authorrunning{}

\author{
Ji\v r{\' i} Hlad{\accent23 u}vka \inst{1}
\and Bui Thi Mai Phuong \inst{1}
\and \\Richard Ljuhar \inst{2}
\and Davul Ljuhar \inst{2}
\and \\Ana M Rodrigues \inst{3,4,5}
\and Jaime C Branco \inst{3,5}
\and Helena Canh\~ao \inst{3}
}
\authorrunning{Ji\v r{\' i} Hlad{\accent23 u}vka et al.}   %

\institute{%
VRVis Center for Virtual Reality and Visualization, Vienna, Austria
\and
Braincon Handels-GmbH, Vienna, Austria
\and
EpiDoC Unit, Centro de Estudos de Doen\c{c}as Cr\'{o}nicas da NOVA Medical School, Universidade Nova de Lisboa, Portugal
\and 
Rheumatology Research Unit, Instituto de Medicina Molecular, Lisboa, Portugal
\and
Servi\c{c}o de Reumatologia do Hospital Egas Moniz, Centro Hospitalar Lisboa Ocidental, Lisboa, Portugal
}

\maketitle

\begin{abstract}
The relationship between knee osteoarthritis progression and changes in tibial bone structure has long been recognized and various texture descriptors have been proposed to detect early osteoarthritis (OA) from radiographs. This work aims to investigate (1) femoral textures as an OA indicator and (2) the potential of entropy as a computationally efficient alternative to established texture descriptors. 
We design a robust semi-automatically placed layout for regions of interest (ROI), compute the Hurst coefficient and the entropy in each ROI, and employ statistical and machine learning methods to evaluate feature combinations.
Based on $153$ high-resolution radiographs, our results identify medial femur as an effective univariate descriptor, with significance comparable to medial tibia. Entropy is shown to contribute to classification performance. A linear five-feature classifier combining femur, entropic and standard texture descriptors, achieves AUC of $0.85$, outperforming the state-of-the-art by roughly $0.1$.
\keywords{computer-aided diagnosis, quantitative image analysis, X-ray imaging, imaging biomarkers}
\end{abstract}

\section{Introduction}

Osteoarthritis (OA) is a degenerative, slowly developing joint disease, characterized by pain and functional disability. It results from a loss of joint cartilage and changes in the subchondral bone area. OA represents one of the leading causes for long-term pain and disabilities associated with musculoskeletal disorders. Thus, early predictors and ways to observe the progression of this disease are highly demanded. 

Traditional radiographic imaging techniques primarily rely on conventional X-rays as opposed to expensive and time-consuming MRI or CT scans. OA examinations focus on semi-quantitative assessment methods like the Kellgren \& Lawrence Score (KL) and quantitative measurements of joint space width (JSW). Such disease parameters have proven to be highly physician-dependent with subjective interpretation results.

Texture information of the trabecular bone as seen in 2D radiographs represents a promising possibility for evaluating the state of OA in addition to traditional clinical means such as visual and semi-quantitative assessments. Algorithms based on fractal analysis of texture \cite{Lespessailles:1994tg} defined by ROIs in tibia have shown to be capable of identifying differences in trabecular bone structure \cite{Wolski-2010-VOT,Thomson-2015-MICCAI,Janvier-2015-IPTA}.

In this work our interest lies in exploring texture-based indicators of OA risk, with a spatial emphasis on femur and a methodological emphasis on entropy. Our contribution to the state-of-the art is three-fold.  

First, we augment the well-researched proximal tibia by ROIs in the distal femur.
In earlier work, ROIs can be found exclusively in proximal tibia: at the medial side \cite{Thomson-2015-MICCAI}, on both sides \cite{Woloszynski:2012ht,Wolski-2010-VOT} or arranged in a lattice inside the tibia head \cite{Janvier-2015-IPTA}. Femur, on the other hand, has so far gone unnoticed with a recent exception unrelated to OA detection \cite{Sampath-2015-JBM}. %
Second, we adopt Shannon entropy for texture characterization. In radiology, it has hitherto been overlooked as a texture descriptor. Only recent research in dental surgery \cite{Koiacinski:2015gw} concludes that the complexity of the texture corresponds to \emph{mature} trabecular bone formation and that entropy turns to be a satisfactory feature for radiograph analyses. 
In the context of knee OA, entropy has only been used in experiments with 3D digital models \cite{Lowitz:2014cy}. 
Third, we develop a simple and interpretable model.
A small set of ROI-texture pairs is identified and combined into a linear classifier. 

Our enquiry proceeds in three steps: ROI placement, feature extraction, and statistical analysis combined with feature selection by machine learning techniques. The steps are detailed in the following section, after a brief description of the data.

\section{Material and Methods}

274 digital radiographs (%
DX modality,
Dexela Detector, %
75 $\mu m$,
$3072 \times 1944$ pixels,
14 bits%
)
of knee joint were available from a study conducted in Portugal \cite{Ramiro:2010wc}. %
Within this sample men and ethnicity other than Caucasian were underrepresented so we retain only 226 Caucasian females for further analyses.
Distal femur and proximal tibia were manually segmented by a technician, and JSW has been measured.
Radiographs were examined and classified as either control or case by three independent radiologists.
The sample of 226 females was further narrowed down to include only subjects for which the three expert-assigned classes coincided. 

The final sample consists of 153 knee radiographs. These amount to 89 left knees and 64 right knees. Subjects were mostly aged between 43.5 and 81 (5\% and 95\% percentile) with a mean of 63.07 and standard deviation of 13.44. Their BMI ranged from 20.97 to 36.80 (5\% and 95\% percentile) with a mean of 28.39 and standard deviation of 4.70. There were 67 cases and 86 controls in total.
                
\subsection{ROI layout}

\newenvironment{myExplanations}{\begin{description}}{\end{description}}
\newcommand{\myExplanation}[1]{\item[#1]}

We introduce an original layout, driven by two main requirements.

\begin{myExplanations}

\myExplanation{Coverage}
We will focus our investigations on the area close to joint space, where the structural changes are believed to be most apparent. \todoBP{cite}
In contrast to prior work which was restricted to proximal tibia, our additional (and novel) effort is to investigate two regions in condyles of the distal femur.
To cover the tibia head we design the layout to avoid the head of fibula.
Similarly, to cover the femoral condyles we position the ROIs so as not to leak into cartilage or into soft tissues on sides.
\myExplanation{Count and size}
We seek to keep the number of ROIs low. The reason for this is related to the curse of dimensionality and the possibility of overfitting with a large number of descriptors. 
Also, we intend to systematically investigate combinations of ROIs and their descriptors to identify the best performing ones. This is not feasible with a large number of ROIs, as they would induce an exponential increase in the number of combinations. 

Concerning ROI size, we prefer to cover the bones by \emph{larger} ROIs to mitigate the volatility of textural discription due to small ROI offsets. %
Larger ROI descriptors also tend to be more discriminative for late OA as recently indicated \cite{Janvier-2015-IPTA}.

\end{myExplanations}

\begin{figure}
	\centering
	\input{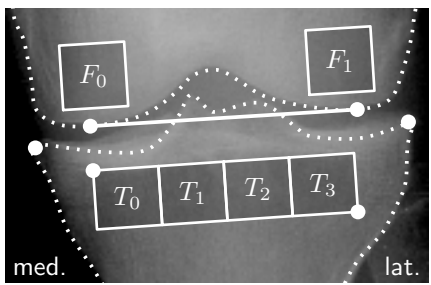}                              %

	\caption{Anchoring ROIs.}
	\label{fig:layout}
\end{figure}

With these requirements in mind we experimentally come up with a layout comprised of six rectangular ROIs, two placed in femoral condyles and four 
in the head of tibia  (cf.~Figure~\ref{fig:layout}).

For consistency across the population we anchor the ROIs at anatomically meaningful landmarks and scale them proporionally. 

\begin{myExplanations}
\myExplanation{Anchoring ROIs in tibia}

The tibia ROIs are determined by medial and lateral tibial plateau landmarks.
Considering a clock-wise unit coordinate system with $(0, 0)$ at medial and $(1, 0)$ at lateral landmarks respectively, 
we span a rectangle with corners at $(0.15, 0.07)$ and $(0.85, 0.23)$.   
Splitting the box into four equal rectangles yields the tibia ROIs $T_0 \dots T_3$ with a height-to-width ratio of $0.9$.  

\myExplanation{Anchoring ROIs in femur}

Contrary to tibia, the femoral ROIs are squares with sides set to the width of the tibia ROIs.
As shape variability of distal femur is larger than that of proximal tibia we need an alternative approach to placement.
The orientation of femoral ROIs $F_0$ and $F_1$ is parallel to the line connecting femoral condyle tips.
Their bottom sides are located 4mm above this line.
To ensure the ROIs do not leak out of the bone they are further horizontally centered within the condyles.

\end{myExplanations}

\subsection{Texture characterization}

The next step is to pick descriptors for the rectangular texture patches defined by ROIs.
For reasons similar to those driving ROI layout, our aim is to keep the number of descriptors low.
From the variety of texture descriptors found in literature we concentrate on two.

First, the Hurst coefficient $\hm\ \in [0,1]$ that appeared in analyses of radiographs in the early nineties \cite{Lespessailles:1994tg},
is related to the fractal dimension of the patch, expresses roughness of the texture and belongs to well established methods in bone research. 

Our second descriptor of choice is Shannon entropy $\en\ = -\sum p_i \log_2 (p_i)$ computed from intensity probabilities $p_i$ within the patch.
Compared to the time-demanding Hurst exponent, computation of entropy is instant.
The theoretical upper and lower entropy bounds for our radiographs are 0 and 14 bits.
The actual measured entropy values within ROIs ranged between 8 and 12.

\subsection{Feature selection: statistical and combinatorial approach}

Six ROIs by two descriptors yield a set of $12$ features for each subject: $\{\hm(F_0),$ $\dots, \hm(T_3), \en(F_0), \dots, \en(T_3)\}$. Along with control-case classification, our data may be represented by a  $153\times 13$ matrix, which is subject to further statistical and combinatorial analyses. We study the importance of each feature, both individually and in the context of other features.

First, the univariate discriminative power of each feature is assessed by means of a two-sample (control vs. case) t-test. The normality assumption is violated for $\hm(F_1)$ and $\en(T_0)$. The results of the respective tests should therefore be interpreted with care, but as will be seen later, they are not central to our discussion.

Second, the interaction between features is taken into account by evaluating each out of $2^{12}-1 = 4095$ nonempty feature combinations. The discriminative ability of each is quantified by the area under the ROC curve (AUC) of a linear support vector machine trained using only the selected features. Estimates of AUC are obtained by averaging over $1000$ five-fold cross-validation runs.

\section{Results}

The p-values from the performed t-tests are shown in Table~\ref{tab:pvalues}. Femur on the medial side turns out to be highly indicative of OA (p-values of $10^{-10}$ and $10^{-7}$ for $\hm(F_0)$ and $\en(F_0)$ respectively). To the best of our knowledge, such findings have not appeared in the literature before.

\setlength\tabcolsep{3mm}
\begin{table}[h]
  \centering
    \caption{p-values from two-sample t-tests} \label{tab:pvalues}
  \begin{tabular}{l | c c c c c c}
        & $F_0$ & $F_1$ & $T_0$ & $T_1$ & $T_2$ & $T_3$ \\ \hline
    \hm\ & $<0.01$ & $0.22$ & $<0.01$ & $0.03$ & $0.01$ & $0.03$ \\
    \en\ & $<0.01$ & $0.27$ & $<0.01$ & $0.63$ & $0.04$ & $0.05$
  \end{tabular}
\end{table}

It is also observed that \en\ exhibits lower significance compared to \hm\ of the same ROI, suggesting that \hm\ is a better univariate descriptor. However, as soon as we enter the multivariate setting, the usefulness of entropy becomes apparent.

\begin{figure}
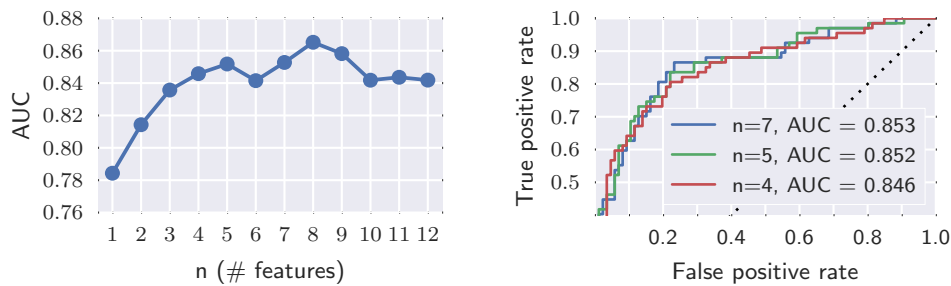

	\centering
	\begin{tabular}{cc}
	\input{auc_by_n_avg.pgf}                              %

	&
	\input{roc.pgf}                              %

	\end{tabular}
	\caption{Left: Best AUCs vs number of features. Right: upper part of selected ROCs.}
	\label{fig:AUCs}
	\label{fig:ROC}
\end{figure}

Let us examine the classification performance that can be achieved by combinations of $n$ features.  
In Figure~\ref{fig:AUCs} we plot for each $n = 1 \dots 12$ the AUC achieved by the best of ${12}\choose{n}$ combinations.

We first note the single-feature AUC of 0.784 -- a value slightly better than results found in past publications on texture-based discrimination of OA ($0.77$ \cite{Woloszynski:2012ht}, $0.754$ in \cite{Thomson-2015-MICCAI}).
Surprisingly, and in contrast to earlier results, this single-ROI descriptor is the medial femur $\hm(F_0)$.   
It should be noted that for $n > 1$ there was always the femoral descriptor $\en(F_0)$ involved in the best subset.  

Our second observation concerns the sudden AUC drop at 6 features. Currently we have a hypothesis rather than evidence regarding its causes: with $n > 5$, the best subsets include descriptors of lateral femur. Its texture, however, was occasionally interleaved with that of patella. 

Ranked by AUC, the top five classifiers are attained for $n = 8,9,7,5$ and $4$ in order. 
The difference in AUC between the best performing 8-subset and the 4-subset is less than 0.02.  
Following the principle of Occam's razor (i.e. to avoid overfitting), we choose to trade the minor gain in AUC for having fewer features. In what follows we concentrate on the three smallest feature sets out of the top five, $n \in \{4,5,7\}$.
Figure~\ref{fig:ROC} shows the corresponding ROC curves. 

\begin{figure}
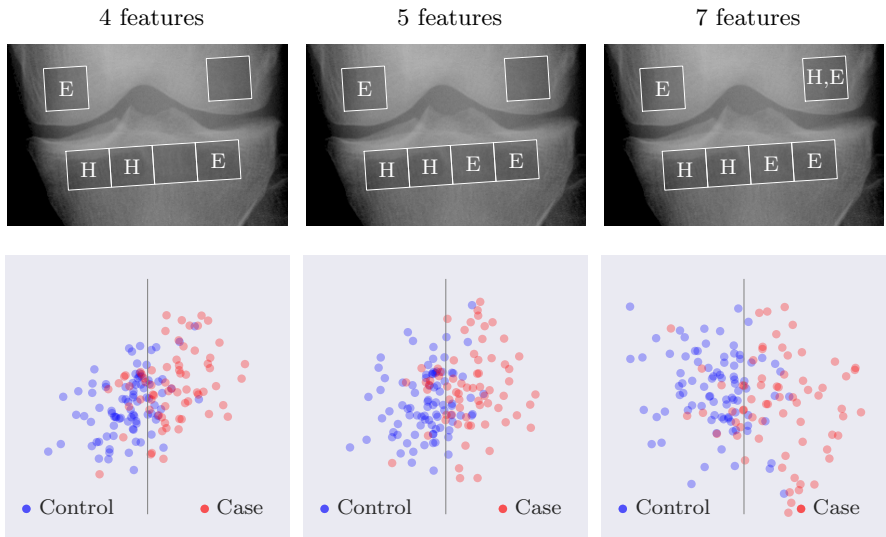

	\centering
	\begin{tabular}{ c c c }
		4 features & 5 features & 7 features
		\\
	\input{selection_04.pgf}                              %
 & 
	\input{selection_05.pgf}                              %
 & 
	\input{selection_07.pgf}                              %

		\\
	\input{projection_04.pgf}                              %
 & 
	\input{projection_05.pgf}                              %
 & 
	\input{projection_07.pgf}                              %
 
	\end{tabular}
	\caption{
		Top row: selected feature set.
		Bottom row: linear, SVM-driven projections of data points from higher-dimensional spaces. 
		Vertical lines represent separation hyperplanes.}
	\label{fig:457}
\end{figure}

We give a graphical overview of the selected feature sets in Figure~\ref{fig:457}.
The four-feature subset describes the medial tibia by \hm\ which is consistent with past findings.
\en\ describes outer lateral tibia $T_3$ and, interestingly, also the medial femur $F_0$ instead of
the missing ROI $T_2$.
The best 5-feature subset fills the tibia gap and, again, \en\ is preferred to \hm.
The best 7 feature subset involves lateral femur described both by \hm\ and \en.
\setlength{\tabcolsep}{-3pt}

From the selection above we point out the 5-feature subset.
Compared to 7 features the difference in AUC is negligible and it needs two fewer features, including the time demanding \hm. 
Compared to 4 features our candidate yields slightly better performance and covers the entire tibia at a negligible additional computational cost.

\begin{figure}
	\centering
	\input{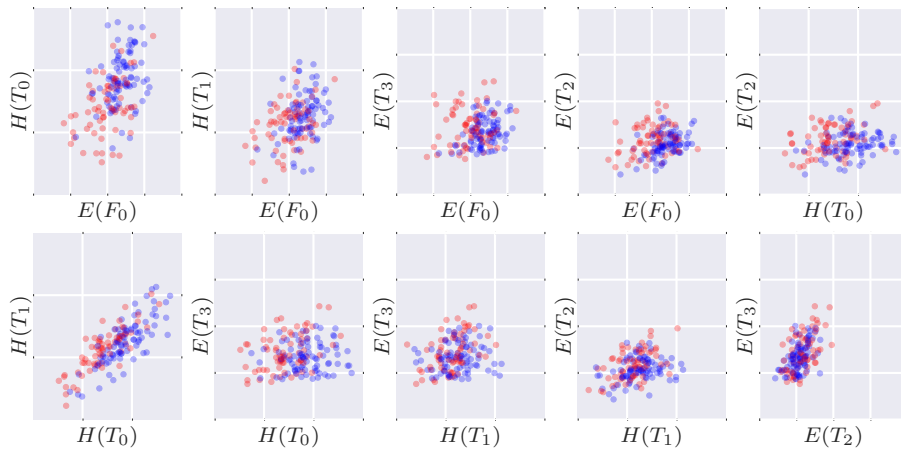}                              %

	\caption{
	Left three columns: all  6 pair plots of the 4-feature space.
	Entire image      : all 10 pair plots of the 5-feature space.
	Grid lines for H(.) at 0.3 and 0.4.
	Grid lines for E(.) at 9, 10, and 11.
	}
	\label{fig:pair_plot}
\end{figure}

The scatter plots in the bottom row of Figure~\ref{fig:457} project data points after normalization to zero mean and unit variance.
The projections are linear and are determined by the discriminating direction of the corresponding SVM and the highest-variance direction of the complementary orthogonal subspace.
Figure~\ref{fig:pair_plot} displays, for $n \in \{4,5\}$, all possible scatter plots with units.

\section{Discussion and Conclusion} 

In this work, we have investigated texture-based descriptors for knee OA detection using a sample of 153 radiographs. In contrast to previous work we examine structural changes in femur and explore the potential of entropy as a texture descriptor. More specifically, entropy \en\ was computed in addition to Hurst exponent \hm\ in each of six semi-automatically placed ROIs, of which four are in tibia and two in femur. Our insights from the study are threefold. 

First, there is indication of OA affecting the bone structure in femur, in addition to the well-known changes in tibia. Support for this observation is provided by highly significant results of the univariate two-sample (case vs. control) t-test, namely a p-value of $2.48\times 10^{-7}$ for entropy measured in medial femur. Moreover, the best feature-set for a given feature-set size consistently included either entropy in medial femur or \hm\ in medial femur, irrespective of the feature-set size. 

Second, the adoption of entropy as a bone texture descriptor is proposed and tested. When considered independently, entropy features do not outperform \hm\ in terms of discriminative power, as suggested by persistently higher p-values. In the multivariate setting however, one may note that the best-performing feature sets always comprise a combination of \hm\ and \en. Indeed, complementing \hm\ with entropy results in a provably better classifier, and one that is computationally more efficient.

Last but not least, a simple model for the prediction of OA risk is developed, using a mixture of femur, entropic and standard texture descriptors. As tested using $1000$ random splits of $5$-fold cross-validation on the available dataset, our model reaches performance levels of 0.85 in terms of AUC, exceeding the state-of-the-art by roughly 0.1. Most importantly, though, the model is better interpretable than currently popular machine-learning techniques such as random forests or neural networks. It combines a small number (five) of well understood texture descriptors by simple linear weighting. The transparency of the model, along with good predictive properties, hint at its potential practical utility as diagnosis aid.

The texture model can further be improved by incorporating shape descriptors \cite{Thomson-2015-MICCAI}.
Other potential extensions include full automation exploiting recent advances in segmentation \cite{Lindner-2013-MICCAI} and/or landmarking \cite{Donner-2013-MIA}.

A possible limitation of our work is the moderate sample size mitigated by the requirement of  classification consistency across three domain experts. This prevented us from segmenting the population by age, BMI or varus / valgus. We leave such investigations with larger datasets for future work.

\enlargethispage{3cm}        
\bibliographystyle{splncs03}
\bibliography{kneeoa}

\end{document}